\documentclass[sigconf,nonacm]{acmart}
\AtBeginDocument{%
  \providecommand\BibTeX{{%
    \normalfont B\kern-0.5em{\scshape i\kern-0.25em b}\kern-0.8em\TeX}}}

\begin{document}

\title{
A Gesture-Based Visual Learning Model for Acoustophoretic Interactions using a Swarm of AcoustoBots
}

\author{Alex Lin}
\affiliation{
\department{Department of Computer Science}
\institution{University College London}
\city{London}
\country{United Kingdom}}

\author{Lei Gao}
\affiliation{
\department{Department of Computer Science}
\institution{University College London}
\city{London}
\country{United Kingdom}}

\author{Narsimlu Kemsaram}
\affiliation{
\department{Department of Artificial Intelligence}
\institution{University of Malaya}
\city{Kuala Lumpur}
\country{Malaysia}}

\author{Sriram Subramanian}
\affiliation{
\department{Department of Computer Science}
\institution{University College London}
\city{London}
\country{United Kingdom}}


\begin{abstract}


AcoustoBots are mobile acoustophoretic robots capable of delivering mid-air haptics, directional audio, and acoustic levitation, but existing implementations rely on scripted commands and lack an intuitive interface for real-time human control. This work presents a gesture-based visual learning framework for contactless human-swarm interaction with a multimodal AcoustoBot platform.
The system combines ESP32-CAM gesture capture, PhaseSpace motion tracking, centralized processing, and an OpenCLIP-based visual learning model (VLM) with linear probing to classify three hand gestures and map them to haptics, audio, and levitation modalities.
Validation accuracy improved from about 67\% with a small dataset to nearly 98\% with the largest dataset. In integrated experiments with two AcoustoBots, the system achieved an overall gesture-to-modality switching accuracy of 87.8\% across 90 trials, with an average end-to-end latency of 3.95 ± 0.43 seconds.
These results demonstrate the feasibility of using a vision-language-model-based gesture interface for multimodal human-swarm interaction. While the current system is limited by centralized processing, a static gesture set, and controlled-environment evaluation, it establishes a foundation for more expressive, scalable, and accessible swarm robotic interfaces.
  
\end{abstract}



\begin{teaserfigure}
  \includegraphics[width=\textwidth]{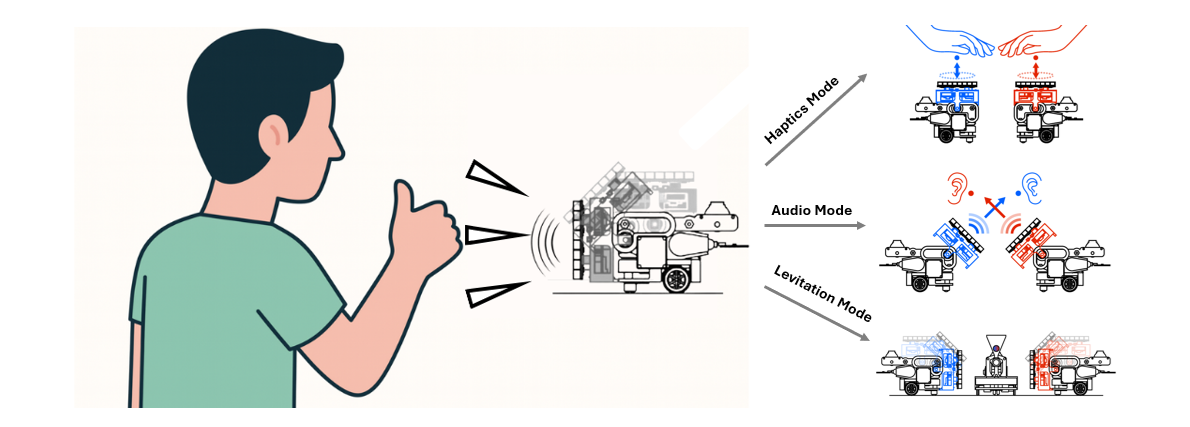}
  \caption{
    Conceptual illustration of the proposed gesture-based AcoustoBot swarm system, where recognized hand gestures are mapped to three multimodal interaction modes: haptics, audio, and levitation.
  }
  \Description{
  }
  \label{teaser}
\end{teaserfigure}

\maketitle

\section{Introduction}

The growing field of multi-agent systems emphasizes the development of swarm robotics that can coordinate to perform complex tasks in a distributed and adaptive manner. Swarm systems, inspired by biological collectives such as ant colonies or bird flocks, are characterized by local sensing, decentralized control, and emerging global behavior. Unlike traditional robotic systems that rely on explicit planning, swarm robotic systems are designed to operate using local rules, producing intricate global behaviors from simple agent interactions \cite{hamann2018swarm}. These characteristics enable the development of new robotics platforms with greater fault tolerance, scalability, and adaptability---traits that are vital for robotics applications in dynamic, real-world environments.
However, as swarm robotic platforms evolve beyond research prototypes into operational tools, a significant constraint remains in this field: How can humans intuitively interact with a collective of autonomous agents in real-time \cite{kolling2015human}? Traditional approaches to human-swarm interaction often rely on abstract command languages, low-level input devices, or pre-scripted behaviors, which make them inconvenient for non-expert users and impractical in fast-paced or dynamic environments \cite{kim2020user}. An effective human-swarm interaction system must strike a balance between control management and system autonomy, enabling the development of high-level abstraction layers that can translate human intent into swarm behavior. 

Recent advancements in VLM provide a pathway toward building semantically aware perception systems. VLMs combine the representational power of deep convolutional neural networks (deep CNNs) with the interpretability of natural language models, allowing machines to not only ``see” but also to ``interpret” visual content in human-like ways. These models embed images and text into a shared latent space, enabling zero-shot classification where a system can infer new classes without task-specific training. 
Trained on large and diverse image–text datasets, VLMs generalize across users and environments, making them well-suited for dynamic and open-ended interaction scenarios such as human-swarm interfaces.
Although most VLM applications have focused on single-robot tasks such as manipulation, navigation, or embodied question answering, recent work highlights their potential in swarm robotics. Unlike deep CNN-based models that rely on fixed-labeled datasets, VLMs flexibly interpret semantic intent expressed through language, systems such as SwarmVLM \cite{zafar2025swarmvlm} and ImpedanceGPT \cite{batool2025impedancegpt} demonstrate how VLMs can modulate swarm behaviors from descriptive input. However, these approaches remain dependent on text-based commands, which, while expressive, may be impractical in many deployment contexts and lack the immediacy of natural, non-verbal interaction.

This paper presents a gesture-based VLM for intuitive human-swarm interaction with a multimodal AcoustoBot platform, as illustrated in Figure \ref{teaser}. The proposed system enables users to issue natural, contactless hand gestures that are interpreted semantically and translated into coordinated robotic behaviors across haptic, audio, and levitation modalities. Rather than relying on conventional interfaces such as keyboards, buttons, or scripted commands, the approach explores how vision-language models can support more accessible and non-verbal interaction with embodied swarm systems. In this way, the work contributes a proof-of-feasibility framework that connects human intent, visual perception, and multimodal robotic actuation in a unified interaction pipeline.

The remainder of this paper is organized as follows: Section 2 reviews related work on hand gesture recognition, vision-language models, and human interaction with robot swarms. Section 3 presents the proposed AcoustoBot system and its gesture-to-modality interaction framework. Section 4 describes the VLM, including feature extraction, linear probing, dataset construction, and deployment. Section 5 reports the experimental evaluation and discusses the results, limitations, and future research directions. Finally, Section 6 concludes the paper.

\section{Literature Review}

This section provides a review of the literature in three areas that underpin this work: (i) multi-agent systems and swarm robotics, (ii) acoustophoresis and multimodal swarm interactions, and (iii) VLMs and human-robot interactions. These studies establish the context and highlight the research gap that this paper aims to address.

\subsection{Multi-Agent Systems and Swarm Robotics}

Swarm robotics focuses on coordinating a large number of simple autonomous robots that communicate and interact locally to exhibit emergent, collective behaviors. Ichihashi et al. introduced Swarm Body \cite{ichihashi2024swarm}, which explores embodied swarm interaction. Unlike traditional multi-agent systems that focus on abstract collective goals, Swarm Body executes in direct, tangible movement with swarm formations. Their experiments demonstrate that users can manipulate swarm configurations in real-time using spatial and gestural cues, effectively treating the swarm as a physical interface and transforming it from an abstract distributed system into direct physical interaction with human users. However, these systems lack seamless input for dynamic human control. This work extends their findings by applying camera-based hand gesture recognition with a VLM. This allows users to control swarm modalities in real-time through semantically grounded, contactless gestures.
Recent works such as SwarmPaint \cite{Serpiva2021SwarmPaint} and Gesture-Controlled Aerial Robot Formation \cite{kratky2025gesture} further demonstrate that gesture-based swarm control can enable intuitive formation and trajectory control, particularly in dynamic and safety-critical environments. These studies help to situate the present work among approaches that provide more expressive interaction between users and robot swarms.

Although decentralization is often regarded as fundamental, this work adopts a centralized strategy due to the limited computational capacity of the AcoustoBots’ onboard microcontrollers. Offloading gesture recognition to a central server ensures real-time performance and reliable coordination. While this departs from fully distributed autonomy, it offers a practical compromise for achieving vision-based gesture interaction. Future work may revisit decentralized approaches as embedded hardware advances.

\subsection{Acoustophoretic Multimodal Interactions}

Acoustophoresis uses acoustic radiation pressure from sound waves to suspend and manipulate particles, cells, or liquids. 
By adjusting phase and amplitude, acoustic traps can be generated and moved, enabling precise, contactless manipulation \cite{becsevli2025sonarios}.
A major milestone was the introduction of GPU-accelerated phase retrieval algorithms, notably GS-PAT (Gerchberg Saxton - Phased Array of Transducers) by Martínez Plasencia et al. \cite{plasencia2020gs}. 
The AcoustoBots platform introduced by Kemsaram et al. integrates mobile ultrasonic phased arrays into robots capable of mid-air haptics, directional audio, and levitation \cite{kemsaram2025acoustobots}. By manipulating transducer phase and amplitude, robots produce acoustic fields that can be felt, heard, or used for levitation \cite{kemsaram2025cooperative}.
%
%
AcoustoBots are modular and scalable, supporting dynamic reconfiguration of formations and coordinated behaviors such as synchronized levitation or multi-user interactions. However, current implementations lack an intuitive control interface. User input is limited to scripted commands or pre-programmed configurations, restricting adaptability. 

To address this gap, this work introduces a vision-based gesture recognition interface powered by CLIP. Using an ESP32-CAM for real-time video, hand gestures are classified into modality-switching commands, transforming AcoustoBots from a programmable platform into a responsive multimodal system for contactless human-swarm interaction.

\subsection{VLM-based Human-Robot Interactions}

Conventional image classifier, often built on CNNs, perform well on curated datasets such as ImageNet but struggle to generalize to new categories without retraining. They tend to memorize training sets rather than learn robust features \cite{zhang2016understanding}, and performance drops significantly when faced with rare classes or diverse real-world environments. This lack of flexibility limits their deployment in human-robot interactions (HRI), where images are often variable, context-dependent, and difficult to capture with fixed labels.
VLMs address these challenges by embedding images and text into a shared space for cross-modal understanding. CLIP, introduced by Radford et al. \cite{radford2021learning}, was trained on 400 million image–text pairs using a contrastive learning objective, enabling zero-shot classification where an image can be matched with the most relevant text prompt without retraining. OpenCLIP \cite{cherti2023reproducible} extended this approach with open datasets, improving scalability and flexibility. These capabilities make VLMs well suited for gesture recognition in HRI, where commands must be interpreted flexibly and in real time.
Gesture recognition in HRI enables a natural, contactless communication channel. Vision-based methods have included skin-colour segmentation, contour modeling, and deep learning, but most approaches face challenges with lighting, background clutter, and generalization \cite{oudah2020hand}. Prior work also emphasizes the lack of a standardized gesture vocabulary, as uninstructed users produce highly variable gestures \cite{tan2021proposed}. 

To address this, this paper integrates OpenCLIP with a predefined set of three static gestures: thumbs up, fist, and palm, mapped directly to AcoustoBot modalities. This approach combines the semantic flexibility of VLMs with a consistent gesture set, enabling intuitive and adaptive multimodal swarm control.

\section{Proposed System}

\begin{figure}[!htbp]
    \centering
    \includegraphics[width=0.45\textwidth]{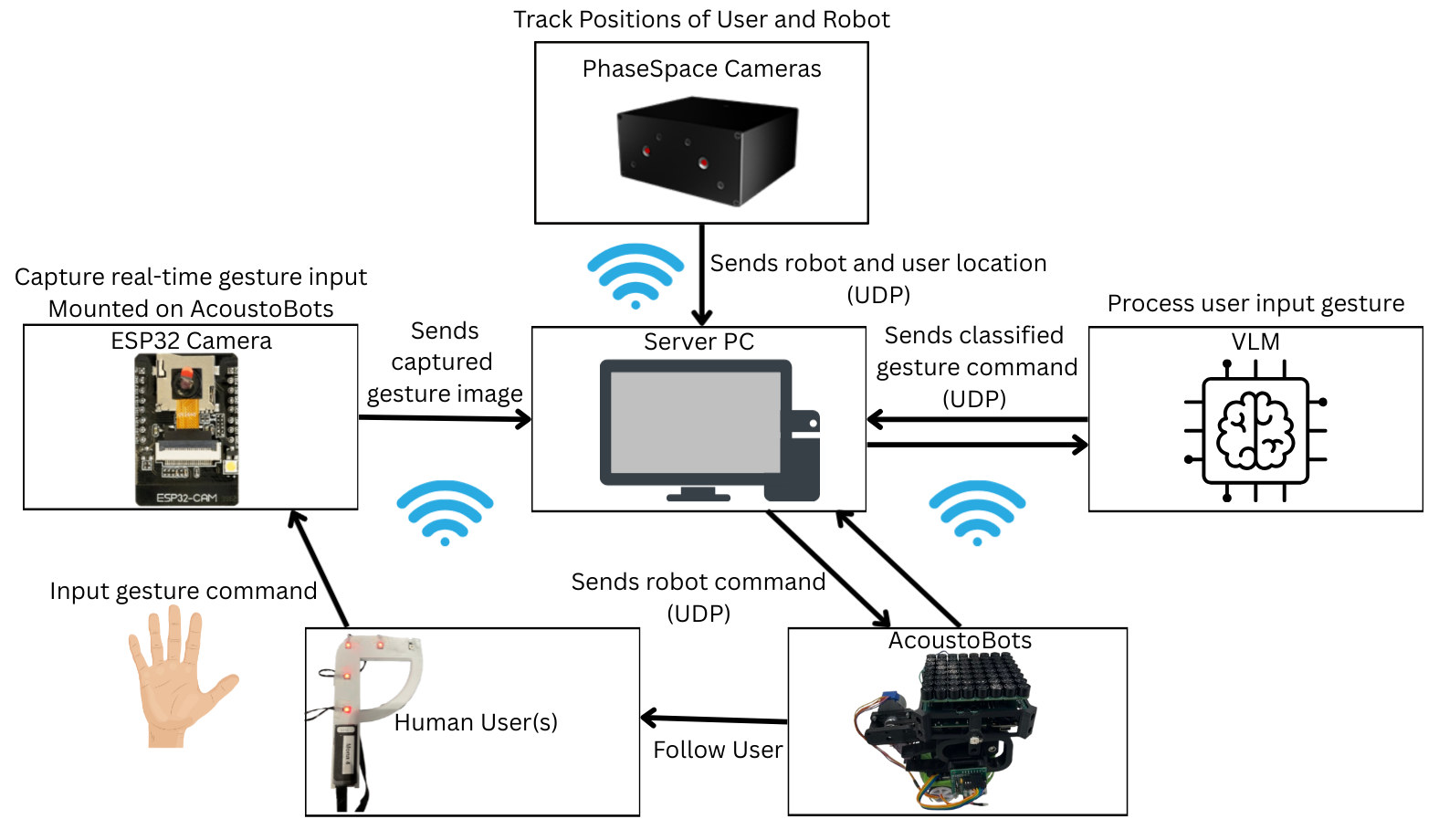}
    \caption{
    System architecture and data flow of the proposed gesture-based human–swarm interaction, showing gesture capture, server-side visual processing, motion-tracking feedback, and command transmission to the AcoustoBots.
    }
    \label{system-diagram}
\end{figure}

This section outlines the architecture of the gesture-based AcoustoBot system, illustrated in Figure \ref{system-diagram}. The system enables real-time, contactless interaction by pairing each robot with a user, responding to both position and gesture commands. It integrates ESP32-CAM modules, a PhaseSpace motion tracking system, a server PC, an OpenCLIP-based gesture VLM, and the AcoustoBot platform.

\subsection{PhaseSpace Motion Tracking System}

To enable robots to follow users autonomously, the system employs the PhaseSpace Impulse X2 motion capture system for precise real-time tracking. LED markers mounted on each AcoustoBot chassis and on user-held handles are captured by PhaseSpace cameras, providing six degrees of freedom position and orientation data. This ensures each user is uniquely associated with a specific robot.
The system continuously streams position data to the central server via UDP, where control algorithms compute heading, speed, and distance adjustments for each robot. This position-feedback loop enables robots to follow their designated users in real time while avoiding collisions.

\subsection{ESP32-CAM for Gesture Input Capture}

Each AcoustoBot is equipped with an ESP32-CAM, a microcontroller with a low-resolution camera, to capture real-time hand gestures for contactless input. This approach creates a low-cost modality control interface without additional sensors or controllers. Distributing cameras across the swarm allows scalable multi-robot interaction.
Captured frames are transmitted via UDP to the central server, where they are preprocessed and forwarded to the VLM. This setup ensures that the robots remain lightweight, with only the necessary sensing hardware on board, while the computationally intensive tasks are handled centrally. Centralized processing reduces robot-side computation and ensures recognition accuracy, two crucial requirements for interactive human-swarm robotics. However, this design choice departs from the ideal of fully decentralized swarm autonomy. Onboard inference is infeasible as the VLM requires significant processing power and memory that far exceeds the capabilities of a simple ESP32 microprocessor. As a result, the system prioritizes functionality and feasibility over decentralization.

\subsection{VLM-based Gesture Classification}

The VLM bridges user intention and robotic actuation by interpreting visual input from the ESP32-CAM and transforming it into functional commands for the AcoustoBots. This allows the system to operate as an intuitive, contactless interface where gestures directly control robot behavior.
Video frames captured by the ESP32-CAM are transmitted to the server, which forwards them to the VLM for classification into predefined gestures. Each gesture is converted into a discrete command, such as ``haptics mode,'' ``levitation mode,'' or ``audio mode.'' The server then integrates these commands with positional data from the PhaseSpace system to ensure that robots execute the appropriate response in the correct spatial context.
In this system architecture, the VLM serves as the interpretation layer between human users and the swarm robots. By converting visual input into semantic commands, it reduces complexity and ensures that user intent is accurately conveyed, supporting scalable and natural human-swarm interaction.

\subsection{AcoustoBot Multimodal Actuation}

At the physical layer, AcoustoBots execute commands via ultrasonic phased array of transducers mounted on motorized hinges, projecting acoustic fields in three-dimensional space. Gestures map to three distinct modalities: an open palm activates haptic feedback, a closed fist triggers directional audio, and a thumbs-up initiates levitation.
This multimodal actuation demonstrates that gesture-based inputs can extend beyond navigation into functional control of swarm modalities. While centralized processing ensures coordination, it also highlights the limitation of reduced autonomy at the individual robot level. Nevertheless, the system proves the feasibility of integrating multiple modalities in a scalable swarm setting, with performance measured through modality switching accuracy, response latency, and user feedback.

\section{Visual Learning Model}


This section presents the VLM that forms the core of the proposed gesture-controlled AcoustoBot system. The model interprets visual input from the ESP32-CAM and maps hand gestures into semantic commands that drive robot behavior. Built on OpenCLIP, the approach leverages pre-trained vision-language representations and adapts them through linear probing for gesture recognition with limited task-specific training data. This design was chosen to preserve transferable visual knowledge, reduce training complexity, and provide a flexible foundation for extending the interaction vocabulary in human-swarm interaction scenarios. The following subsections describe the model architecture, feature extraction pipeline, dataset and training methodology, deployment process, and integration with AcoustoBot control.

\subsection{OpenCLIP Architecture}

Our VLM is built on OpenCLIP, an open-source implementation of CLIP developed by OpenAI. OpenCLIP has been pre-trained on large-scale image–text datasets, enabling it to learn semantically meaningful representations across domains. The model employs a Vision Transformer (ViT) backbone \cite{dosovitskiy2020image}, which divides each image into non-overlapping patches embedded in a sequence processed by transformer layers. This allows the model to capture both local features, such as finger orientation, and global context, such as overall hand pose.
Pre-training employs contrastive learning, aligning images and text in a shared embedding space \cite{chen2020simple}. Matching pairs are positioned closer together, while unrelated pairs are placed further apart. The visual encoder produces high-dimensional embeddings that capture semantically meaningful features. By mapping gesture images into the same representation space as natural language labels, the model provides a flexible mechanism for gesture classification and supports future extension to richer interaction vocabularies without redesigning the entire perception pipeline.

\subsection{Feature Extraction and Linear Probing}

OpenCLIP is adapted for gesture recognition through a two-stage process. First, the pre-trained visual encoder is frozen and used as a feature extractor. The input images are resized to 224×224, normalized according to ImageNet statistics, and converted to the tensor format \cite{he2016deep}. Each frame is encoded into a 512-dimensional vector that captures the semantic characteristics of the gesture \cite{zhou2022learning}.
In the second stage, a lightweight linear probe maps these embeddings into three gesture categories: palm, fist, and thumbs up. This classifier is implemented as a single fully connected layer, trained while the encoder remains fixed. Linear probing preserves general visual knowledge, avoids catastrophic forgetting \cite{kirkpatrick2017overcoming}, reduces trainable parameters \cite{qiao2020learning}, and improves efficiency with small datasets \cite{kumar2022fine}.

\subsection{Dataset Construction and Training}

The training dataset comprises three categories: thumbs up, fist, palm, sourced from both lab-captured images and public databases. It includes variations in lighting, background, and hand orientation to improve generalization. The annotation process involves labeling each image with its corresponding gesture class, creating a tab-separated annotation file that maps image filenames to their labels. The training process of the VLM follows a supervised learning paradigm where the system learns to associate visual features with discrete gesture labels. The dataset is divided into an 80/20 train-validation split, where 80\% of the data is used to train the model and the remaining 20\% is reserved for validation. This setup ensures that the model performance is regularly assessed on unseen data, reducing the risk of overfitting.
The optimization process employs the AdamW algorithm with a learning rate of 1e-3. AdamW optimizer combines the adaptive learning rate adjustment of the original Adam algorithm with decoupled weight decay, which helps to control overfitting and improve convergence stability \cite{loshchilov2017decoupled}. The model uses Cross-Entropy Loss as the objective function, a standard approach for multi-class classification problems that provides stable gradients and reliable feedback during training \cite{goodfellow2016deep}.
Training was carried out over 50 epochs with a batch size of 5, chosen to strike a balance between computational efficiency and gradient stability. Although larger batch sizes may speed up training, they often lead to poorer generalization. In this instance, a smaller batch size is appropriate for the dataset size and helps prevent overfitting while maintaining stable training dynamics \cite{keskar2016large}. During each epoch, the training data is passed through the network, gradients are computed, and the parameters of the linear probe are updated, while the CLIP encoder remains frozen. This design ensures that the model benefits from the pre-trained features of CLIP while efficiently adapting to the specialized task of gesture recognition.

\subsection{Model Exportation and Deployment}

After training, the linear probe was saved in PyTorch format with class mappings for consistency. For real-time gesture inference, images are preprocessed, passed through the frozen CLIP encoder, and classified by the linear probe model, which outputs a set of raw values, known as logits. Logits can be understood as the normalized confidence score of the classification model, and they may be positive or negative and do not have any probabilistic meaning yet. Each gesture class has its own corresponding logit. The logits are converted to probabilities using the softmax function \cite{bridle1989training}, and the highest-probability prediction is selected.
For deployment, the trained model is exported to ONNX (Open Neural Network Exchange) format \cite{bai2019onnx}, combining the encoder and probe into a single computational graph. This integration avoids the need for separate model loading or feature extraction steps during deployment. By adopting ONNX, the system achieves platform independence and compatibility with various deployment frameworks. Notably, ONNX exportation enables the model to run seamlessly on Windows platforms, where the AcoustoBot application is executed. This enables efficient inference on the central server, optimizes runtime performance, and simplifies integration with the AcoustoBot control pipeline.

\subsection{Integration with AcoustoBot Control}

\begin{figure}[!htbp]
    \centering
    \includegraphics[width=0.45\textwidth]{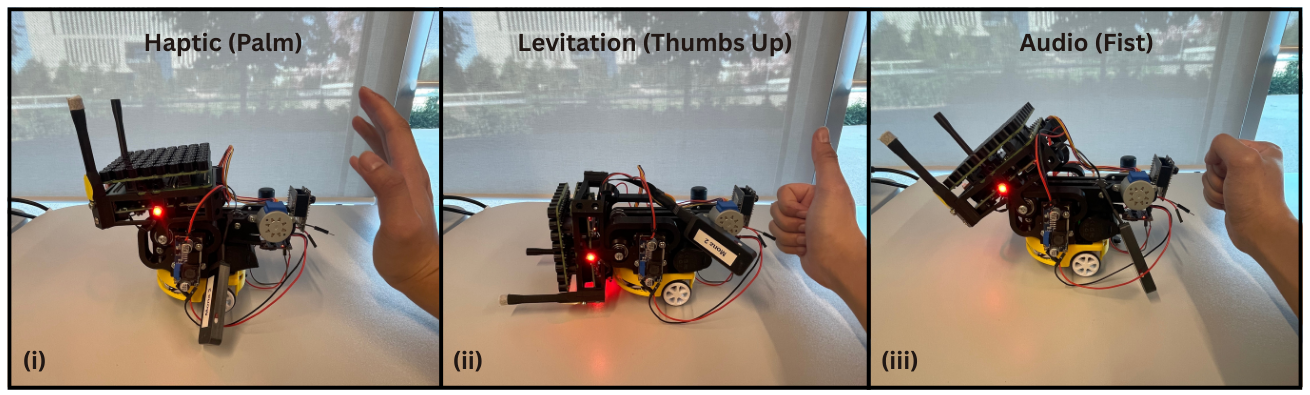}
    \caption{
    Gesture-to-modality mapping in the proposed AcoustoBot interaction framework. Open palm activates haptic mode, thumbs up activates levitation mode, and fist activates audio mode. Each gesture is captured by an ESP32-CAM and interpreted by the vision-language model to trigger the corresponding multimodal robotic response.
    }
    \label{gesture-modality}
\end{figure}

The classification outputs integrate with the AcoustoBot software, mapping gestures to specific modalities: palm activates haptic feedback, fist triggers audio, and thumbs up initiates levitation, as shown in Figure \ref {gesture-modality}. To improve robustness, the system uses confidence scores as a rejection threshold. Low-confidence predictions are discarded to prevent unintended robot actuation. This confidence-based mechanism ensures that robots respond only when the system is sufficiently certain, aligning the actuation with user intent.

\section{Evaluation, Results, and Discussions}

The evaluation is conducted in two parts to assess both the performance of the gesture classification model and its integration within the AcoustoBot platform. The first part focuses on evaluating the VLM, examining metrics such as training loss, validation loss, and classification accuracy. The second part evaluates the end-to-end integration of the model with the AcoustoBot system through experimental setups, measuring response time and the accuracy of modality switching. Together, these evaluations assess both the technical validity of the model and its practical effectiveness for intuitive real-time human-swarm interaction.

\subsection{Evaluation of Gesture Classification Model}

The performance of the gesture classification model is first assessed independently. The dataset was divided into an 80:20 training-to-validation split to analyze generalization capability. Training and validation loss were recorded over 50 epochs to evaluate convergence behavior and potential overfitting. A total of 8 training runs were recorded, each using a different dataset size, to compare how dataset scale influences generalization. Primary metrics were training loss, validation loss, validation accuracy, and training speed was also compared across runs.

\subsubsection{Training Speed}

\begin{figure}[!htbp]
    \centering
    \includegraphics[width=0.30\textwidth]{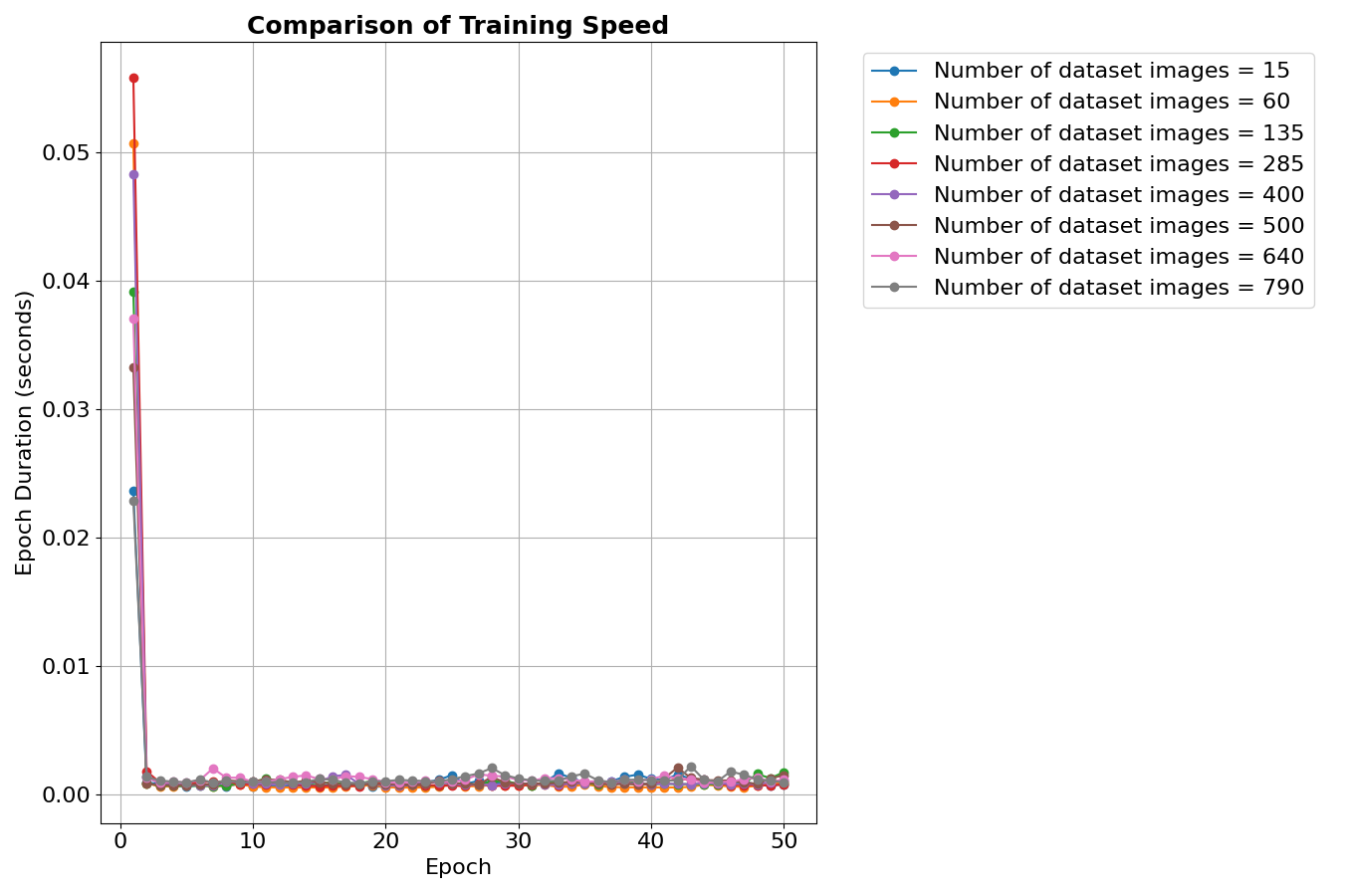}
    \caption{
    Comparison of training loss over epochs for different dataset sizes. The results show fast convergence in all cases, indicating that the linear probing approach remains efficient even as the training set grows.
    }
    \label{training-speed}
\end{figure}

As shown in Figure \ref{training-speed}, epoch duration remained consistently low across all runs, with subsequent epochs converging to millisecond runtime (typically 0.001–0.05 seconds) even as the dataset increased from 15 to 790 images. This demonstrates that the linear probing approach maintains computational efficiency regardless of dataset size.
This efficiency supports rapid retraining or extension (e.g., adding new gesture classes) without expensive computational resources and improves the scalability of the interaction control system.

\subsubsection{Training and Validation Loss}

\begin{figure}[!htbp]
    \centering
    \includegraphics[width=0.30\textwidth]{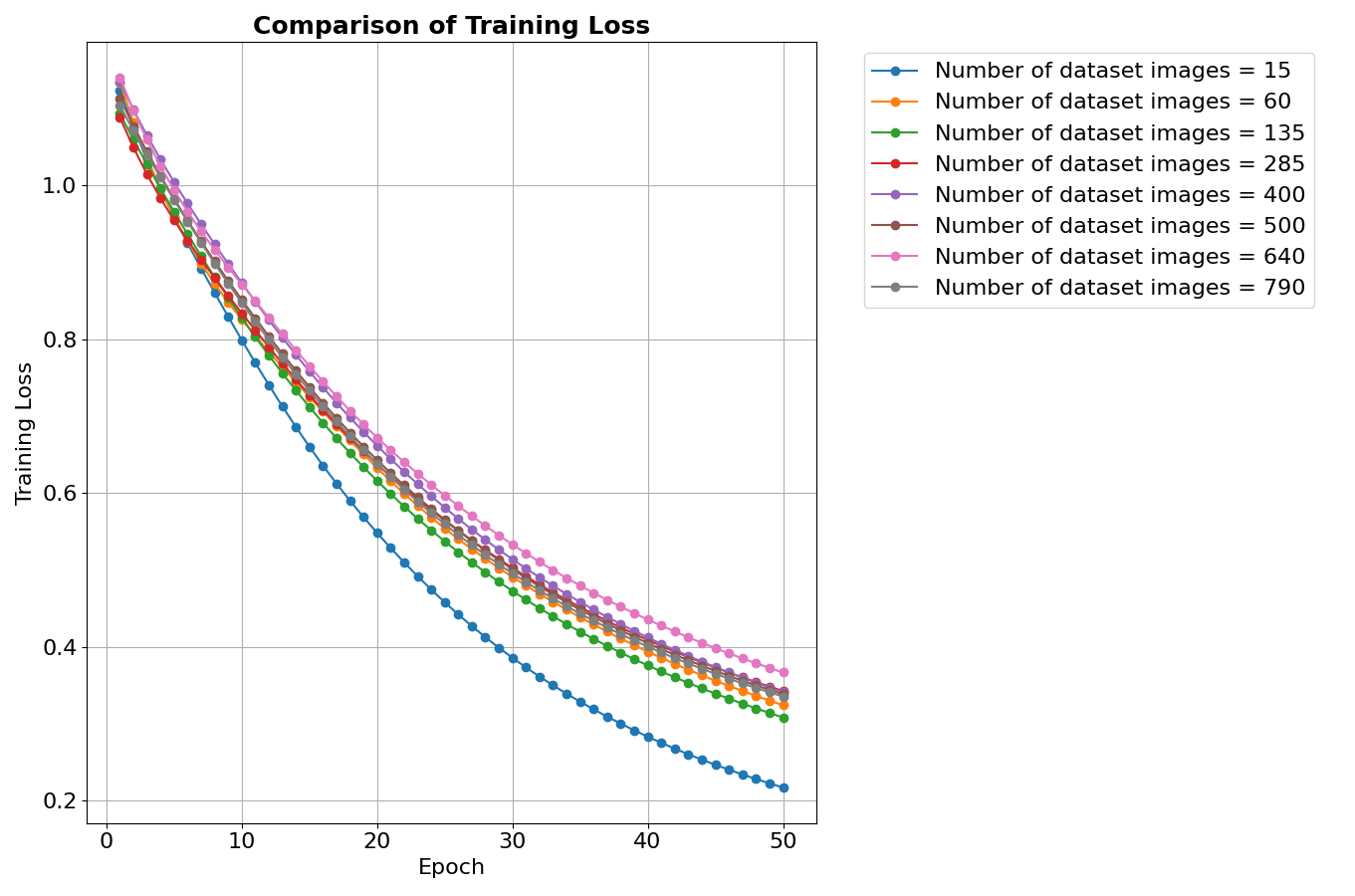}
    \caption{
    Training loss curves for different dataset sizes, showing progressive reduction in loss over epochs and improved learning stability as the amount of training data increases.
    }
    \label{training-loss}
\end{figure}

\begin{figure}[!htbp]
    \centering
    \includegraphics[width=0.30\textwidth]{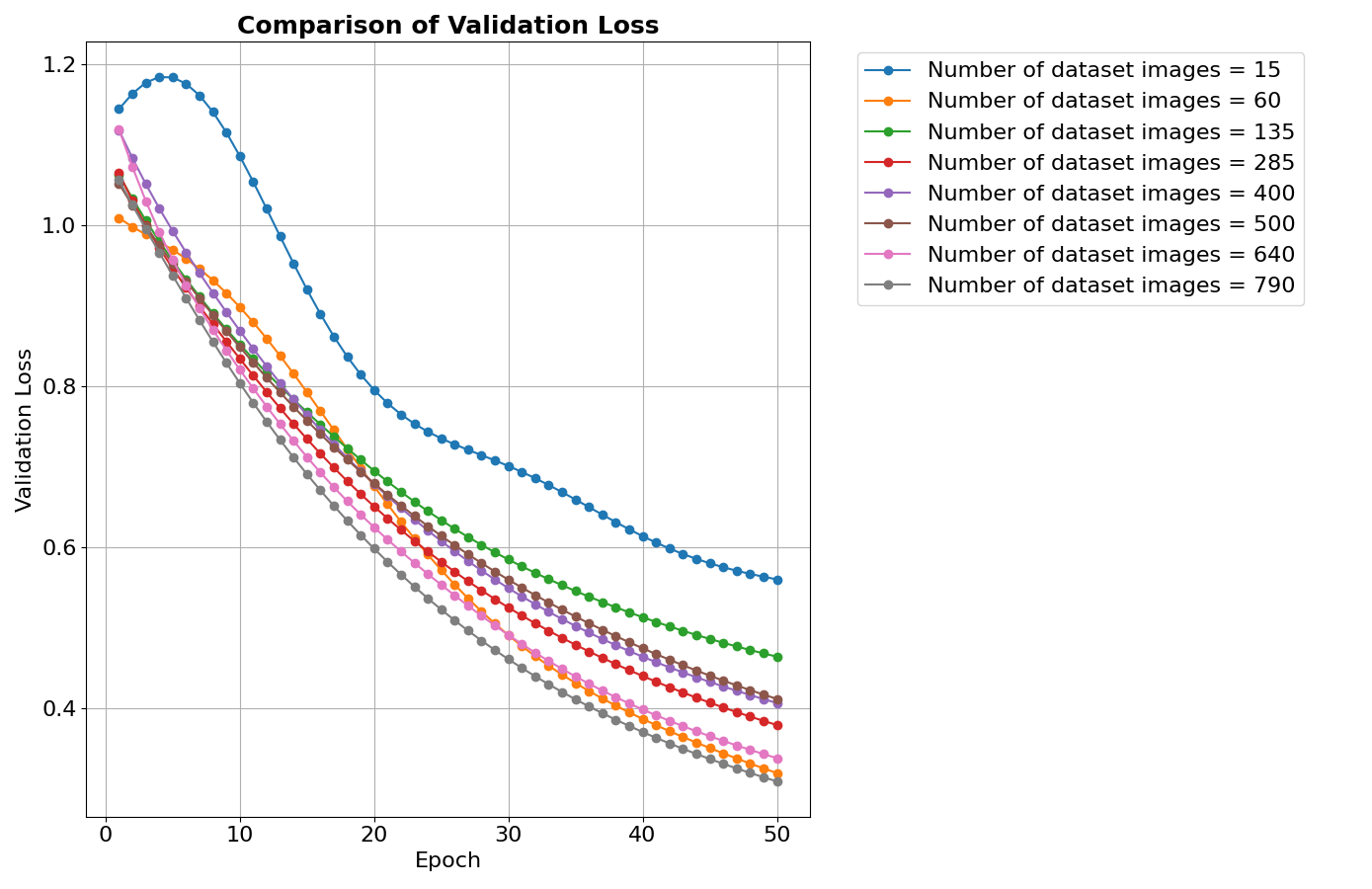}
    \caption{
    Comparison of validation loss over epochs across dataset sizes. Larger datasets yield lower validation loss and smoother convergence, indicating better generalization performance.
    }
    \label{validation-loss}
\end{figure}

The performance of the gesture classification model was evaluated using training and validation loss across progressively larger dataset sizes. Training loss measures how well the model learns to associate gesture images with their labels during training, while validation loss indicates how well the model generalizes to unseen samples. Both are computed with the cross-entropy loss function, where lower values reflect improved learning and generalization.
Figures \ref{training-loss} and \ref{validation-loss} show that loss values improve as dataset size increases. Training loss consistently decreased from initial values around 1.10 to below 0.40 by the 50th epoch. Although the 15-image dataset appeared to converge faster, it is a sign of severe overfitting and poor generalization. The key insight lies in examining the validation loss alongside the training loss. The validation loss of the 15-image dataset initially increased (from 1.14 to 1.18) during the first 10 epochs before decreasing, indicating that the model memorizes the small training set rather than learning generalizable features \cite{qiao2020learning}. 
%
%
%
%
These results confirm that performance benefits substantially from dataset scaling, reinforcing the importance of diverse training data. The decline in validation loss shows that the model can distinguish the three gesture classes: thumbs up, fist, and palm, when provided with sufficiently varied inputs. From a system design perspective, this also suggests scalability: the architecture could support additional gesture classes in the future if paired with adequate training data.

\subsubsection{Validation Accuracy}

\begin{figure}[!htbp]
    \centering
    \includegraphics[width=0.30\textwidth]{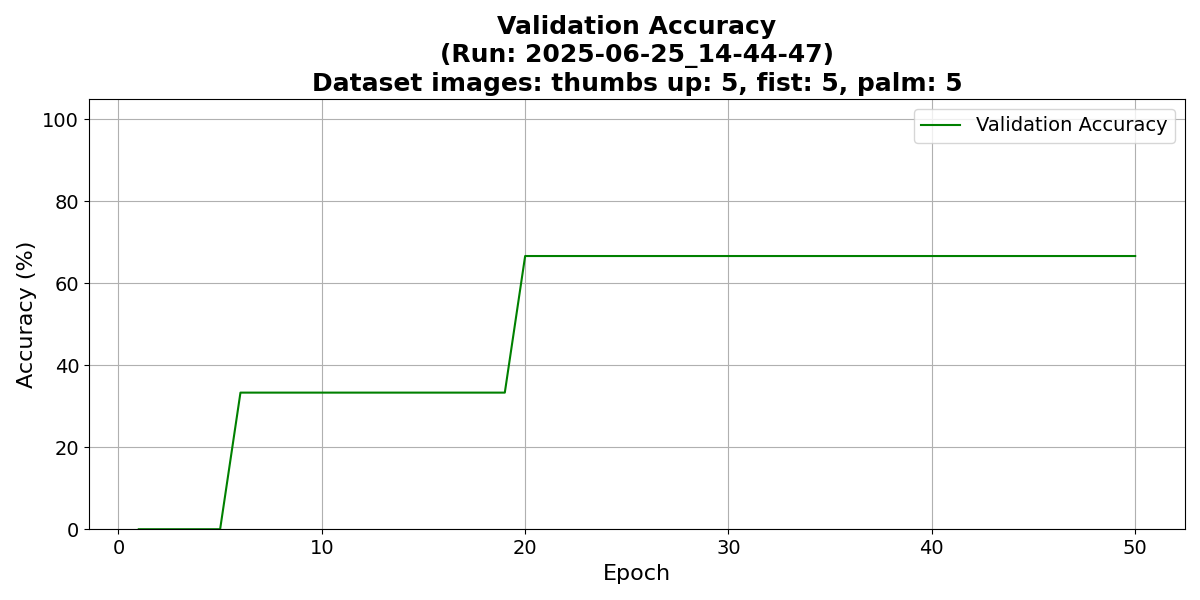}
    \caption{
    Validation accuracy curve for the first training run using 15 images, showing limited performance and early saturation due to the small dataset size.
    }
    \label{accuracy-curve-f}
\end{figure}

\begin{figure}[!htbp]
    \centering
    \includegraphics[width=0.30\textwidth]{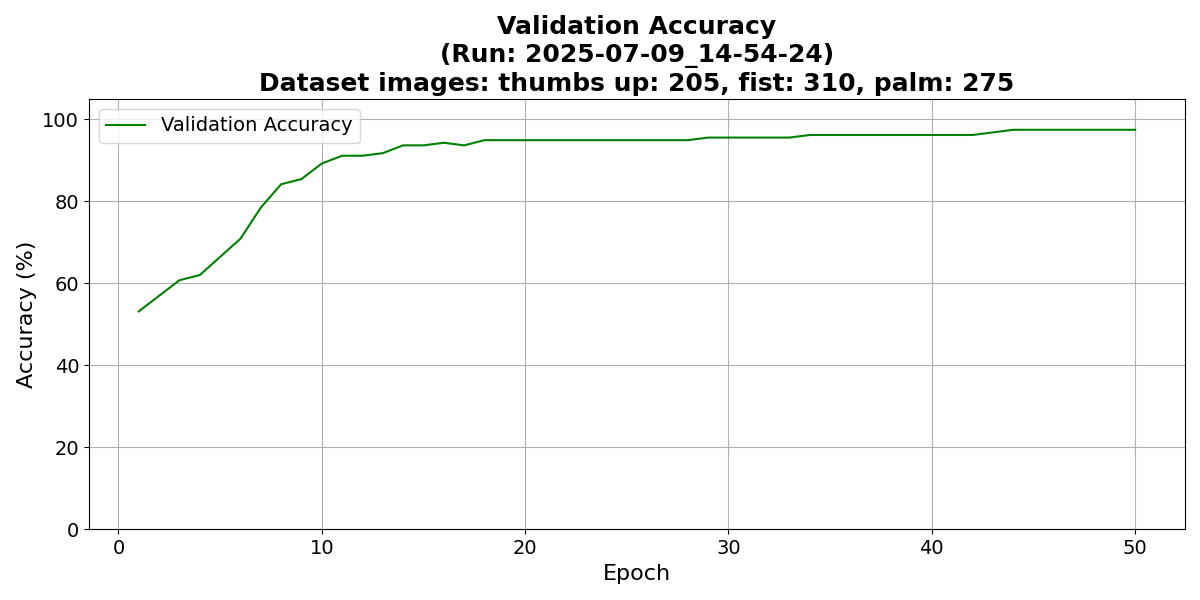}
    \caption{
    Validation accuracy curve for the final training run using 790 images, showing rapid improvement and near-saturated performance as the model benefits from a larger and more diverse dataset.
    }
    \label{accuracy-curve-l}
\end{figure}

In addition to training and validation loss, validation accuracy was recorded after each run to assess generalization to unseen data. Validation accuracy, expressed as the percentage of correctly classified validation samples, provides a direct measure of overall model performance.
The first run, conducted with only 15 images (five per class), achieved a final validation accuracy of about 67\%. As shown in Figure \ref{accuracy-curve-f}, the accuracy curve plateaued after roughly 20 epochs. This suggests that the model quickly memorized the small dataset but could not improve further due to the lack of diversity. With so few samples, the model overfit to the training set, failing to learn generalizable representations of the gesture classes.
By contrast, the most recent run with 790 images demonstrated substantial improvement. Figure \ref{accuracy-curve-l} shows validation accuracy steadily increasing and stabilizing to near 98\%. The larger dataset provided sufficient diversity and complexity for the model to learn robust, transferable patterns, reducing overfitting and supporting reliable classification across varied conditions.
These results emphasize the critical role of dataset size in achieving high recognition performance. Small datasets may lead to rapid convergence but yield limited accuracy and poor generalization. Larger, more diverse datasets enable the model to capture richer gesture features, resulting in higher accuracy and greater robustness when deployed in real-world human–robot interaction scenarios.

\subsection{Evaluation of AcoustoBots}

\subsubsection{Experimental Setup}

\begin{figure*}[!htbp]
    \centering
    \includegraphics[width=0.60\textwidth]{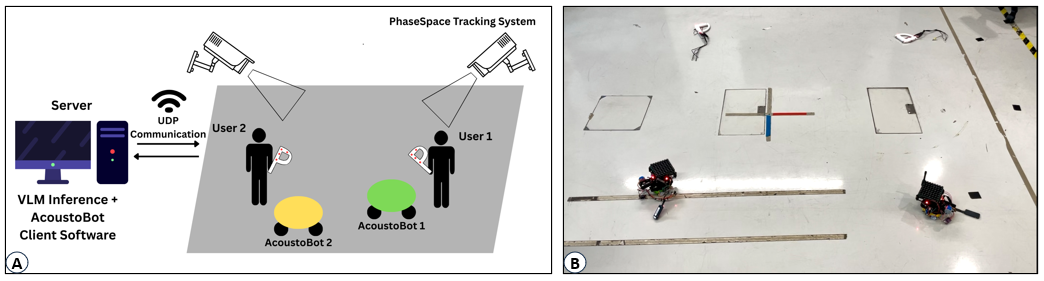}
    \caption{
    Experimental setup for the AcoustoBot gesture-interaction evaluation: A) schematic of the test arena showing the VLM inference server, PhaseSpace tracking system, users, and AcoustoBots, and B) real lab arena used for evaluation.
    }
    \label{fig:experiment-setup}
\end{figure*}

The integration experiment was conducted in a controlled laboratory environment (as shown in Figure \ref{fig:experiment-setup}) with a server PC, two AcoustoBots, two ESP32 cameras, two user handles, and a PhaseSpace tracking system. Each AcoustoBot is equipped with acoustic control client software and an ESP32 camera module to capture live gesture frames; PhaseSpace enabled real-time motion tracking for user-following behavior. Position and gesture streams were transmitted to the server, which computed navigation commands with collision avoidance.
%
%
Collectively, the experiment setup integrated the system's three key data streams: i) AcoustoBot and user position from PhaseSpace, ii) gesture input from ESP32 cameras, and iii) robot actuation feedback, into a coherent framework for demonstrating real-time human-swarm interaction.

\subsubsection{Experimental Result}

\paragraph{Accuracy of Modality Switching}


The results of gesture-to-modality switching are summarized in Table~\ref{tab:modality_accuracy}, which shows an overall accuracy of 87.8\%.
Thumbs up was lower than fist and palm, likely due to finer structural details and ESP32-CAM resolution constraints, which increase sensitivity to hand orientation, background, and lighting. Despite this, the overall accuracy supports real-time multimodal control, the compact gesture vocabulary remains effective and can be improved with higher-quality sensing or augmentation.
\begin{table}[!htbp]
\caption{Gesture-to-modality switching accuracy in the integrated AcoustoBot system.}
\label{tab:modality_accuracy}
\centering
\begin{tabular}{lccc}
\toprule
Gesture Class & Trials & Correct & Accuracy (\%) \\
\midrule
Thumbs Up & 30 & 24 & 83.3 \\
Fist & 30 & 26 & 86.7 \\
Palm & 30 & 29 & 96.7 \\
\midrule
\textbf{Overall} & \textbf{90} & \textbf{79} & \textbf{87.8} \\
\bottomrule
\end{tabular}
\end{table}
Overall, the evaluation demonstrates that the proposed system can reliably interpret a small set of hand gestures and translate them into multimodal robot behaviors in real time under laboratory conditions. Validation accuracy improved substantially as dataset size increased, reaching near 98\% in the largest training run, while integrated system testing achieved an overall gesture-to-modality classification accuracy of 87.8\%. These findings support the feasibility of using a VLM-based perception pipeline for contactless human-swarm interaction, while also indicating that broader gesture vocabularies and more diverse user conditions will require further data collection and optimization.

\paragraph{Response Latency}

\begin{figure}[!htbp]
    \centering
    \includegraphics[width=0.30\textwidth]{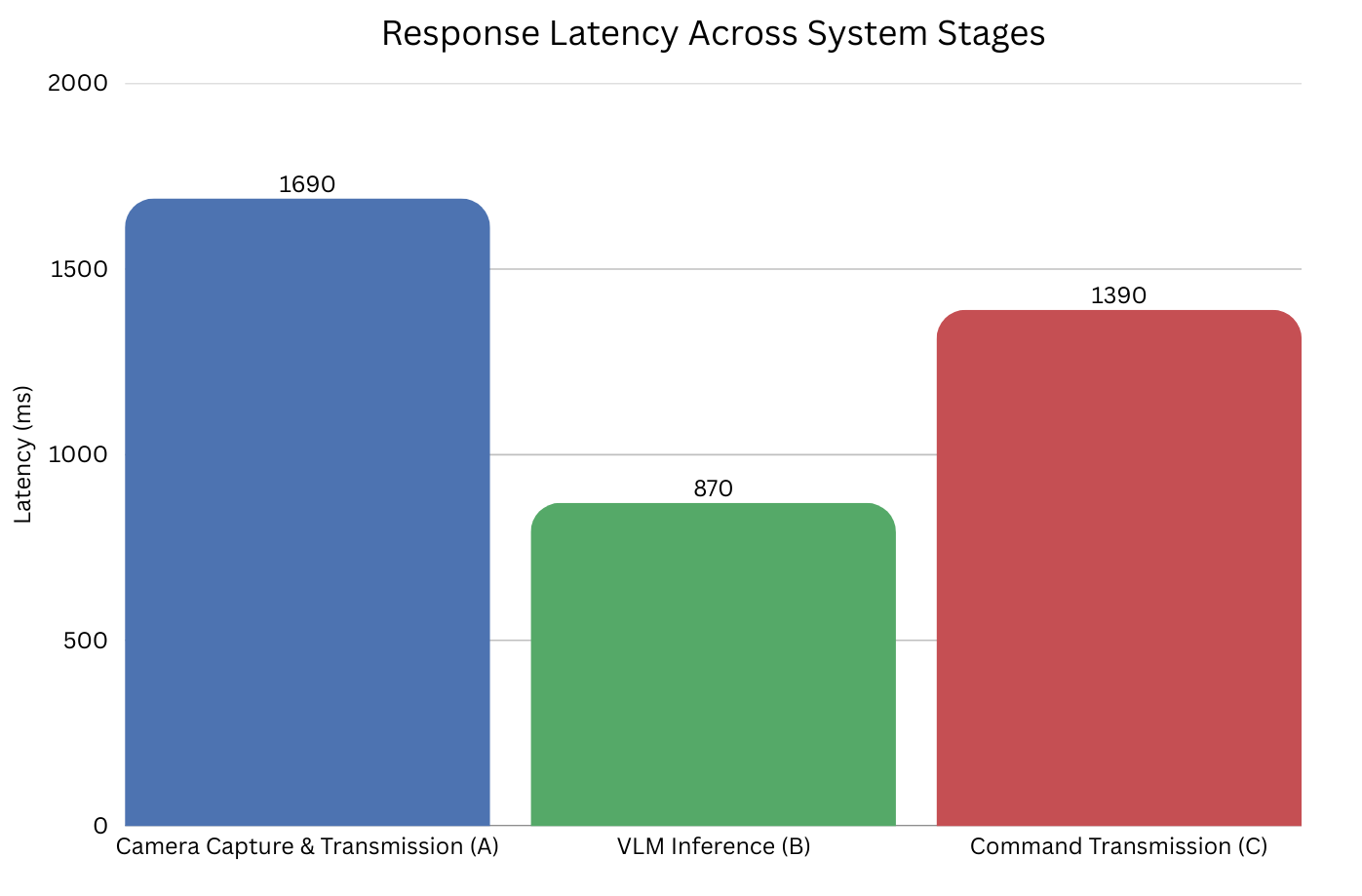}
    \caption{
    Average response latency across the main stages of the proposed system: (A) image capture and transmission from the ESP32-CAM to the central server, (B) VLM-based gesture classification time, and (C) modality command transmission time.
    }
    \label{latency-stage}
\end{figure}

\begin{figure}[!htbp]
    \centering
    \includegraphics[width=0.30\textwidth]{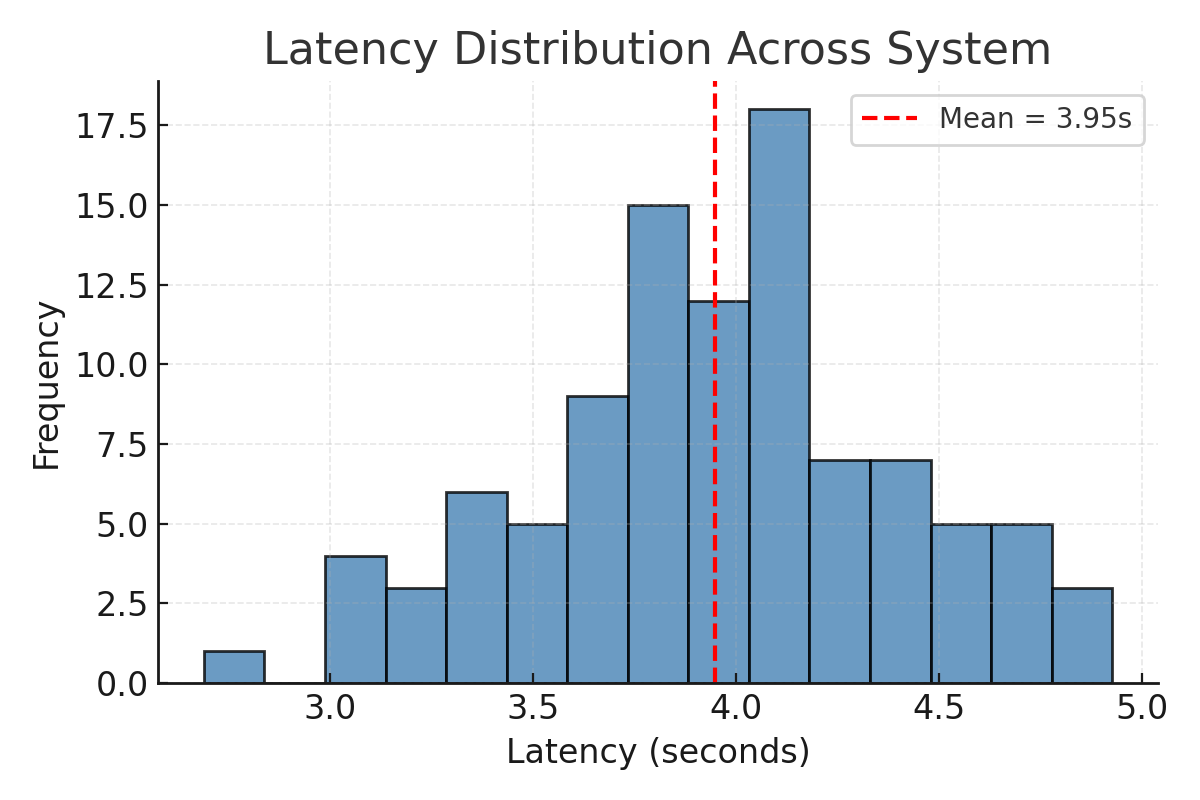}
    \caption{
    Distribution of end-to-end command execution latency across 50 trials, showing the variability of system response times and a mean latency of 3.95 seconds.
    }
    \label{latency-dis}
\end{figure}

Latency was defined as the elapsed time from gesture to robot actuation and recorded via timestamps at each stage. Measurements across 10 trials per modality decomposed total time into image capture/transmission, VLM inference, and command transmission, as presented in Figure \ref{latency-stage}. The majority of  delay arose from camera capture/transmission, which is primarily attributed to the limited hardware capabilities of the ESP32 camera module. VLM inference remained relatively low due to the linear probe design.
%
%
In addition, Figure \ref{latency-dis} shows the distribution of response times, with most commands executed in under 5 seconds, yielding an average latency of 3.95 seconds with a standard deviation of $\pm$0.43 seconds. These results indicate that the system is responsive enough for proof-of-feasibility interaction and multimodal control demonstrations, although further optimization is needed to support smoother real-time use in more demanding settings. Much of the delay arose from image capture and transmission, suggesting that improved onboard imaging hardware and more efficient communication pipelines could substantially enhance responsiveness.

\subsection{Discussion}


This work addresses the challenge of designing a natural and intuitive interface for human interaction with a swarm of multimodal robots. The proposed system uses camera-based gesture recognition powered by a vision-language model to control AcoustoBots through three mapped modalities: open palm for haptics, fist for audio, and thumbs up for levitation. In doing so, the system moves beyond conventional symbolic or device-dependent control schemes and explores how non-verbal visual input can support accessible human-swarm interaction.
Across trials, the system achieved an average classification accuracy of approximately 87.8\% with mean latency below four seconds, demonstrating the feasibility of bridging semantic perception and embodied robotic actuation within a swarm setting. The primary contribution lies not only in the implementation of the perception pipeline, but also in showing how simple human gestures can be translated into meaningful multimodal swarm responses. This positions the system as an initial step toward more expressive, user-friendly interfaces for multi-agent robotic platforms.
This work advances the vision of Ichihashi et al. \cite{ichihashi2024swarm} on embodied swarms responsive to human movement and demonstrates how simple local rules can yield complex behavior when coupled with a carefully designed interaction framework. By integrating haptics, audio, and levitation, the system extends interaction beyond spatial coordination to functional expressivity, showing that gesture semantics can effectively transform user intent into multimodal actuation.
%
Several limitations remain. First, the current system relies on centralized gesture classification on a server, which constrains scalability and introduces latency. Second, the present study focuses on only three static gesture classes and evaluates performance primarily in controlled laboratory conditions. Third, while the system demonstrates technical feasibility, it does not yet include a user study or comparative evaluation against alternative gesture-recognition approaches. Future work should therefore investigate distributed on-robot inference using more capable embedded hardware or compressed VLMs, expand the gesture vocabulary and user diversity, and include formal human-subject evaluation to assess usability, learnability, workload, and interaction quality.
%
Overall, the results demonstrate that simple hand gestures can reliably control complex swarm behaviors, offering a scalable and intuitive framework and pointing toward levitation-enabled swarm robots with decentralized intelligence operating in everyday environments.

\section{Conclusion}


This paper presented a gesture-based visual learning framework for controlling a swarm of multimodal AcoustoBots through natural, non-verbal interaction. By mapping simple hand gestures to haptic, audio, and levitation functions, the system demonstrated how human intent can be interpreted visually and translated into embodied swarm behaviors. The results show that a vision-language-model-based approach can provide a practical proof of feasibility for contactless human-swarm interaction, while also highlighting the need for lower-latency hardware, richer gesture vocabularies, and broader user evaluation. Overall, the work establishes a foundation for more expressive and accessible interfaces for future multimodal swarm robotic systems.

\section{Acknowledgments}

This work was supported by the EPSRC Prosperity Partnership Program - Swarm Spatial Sound Modulators (EP/V037846/1), and by the Royal Academy of Engineering through their Chairs in Emerging Technology Program (CIET 17/18).


\bibliographystyle{unsrt}
\bibliography{acmart}

\appendix

\end{document}